\documentclass{article}
\usepackage{spconf,amsmath,amssymb,graphicx,booktabs,xcolor,url,enumitem,etoolbox,cite}
\usepackage[T1]{fontenc}
\setlist{nosep,leftmargin=14pt}
\AtBeginEnvironment{thebibliography}{\setlength{\itemsep}{0pt}\setlength{\parskip}{0pt}}
\definecolor{todocolor}{RGB}{165,20,70}

\newcommand{\E}{\mathbb{E}}
\newcommand{\mosaicfull}{\textbf{M}odality-aware \textbf{O}rthogonal-axis \textbf{S}tatus-conditioned \textbf{A}ttentive \textbf{I}ntegration for \textbf{C}ompletion}

\title{MOSAIC-Flow: Orthogonal-Axis and Status-Conditioned Fusion for Missing-Sequence Completion in Brain Tumor MRI}

\name{Pengli Zhu \qquad Zhihao Peng \qquad Yixuan Yuan$^{\star}$}
\address{Department of Electronic Engineering,
The Chinese University of Hong Kong, Hong Kong, China \\
$^{\star}$ Corresponding author: Yixuan Yuan}

\begin{document}
\maketitle

\begin{abstract}
\sloppy
Missing-sequence completion is important for comprehensive brain tumor characterization, but cross-modal fusion can obscure sequence-specific contrast and lose fine-grained three-dimensional tumor detail. To address this challenge, we introduce \mosaicfull\ (\textbf{MOSAIC-Flow}), a unified framework for missing-sequence completion that learns a target-conditioned velocity field while keeping the seven acquired volumes fixed throughout the transport. Specifically, modality-aware stems independently encode each sequence, and orthogonal-axis refinement captures axial, coronal, and sagittal dependencies while retaining the three-dimensional representation. Status-conditioned prototype fusion uses compact modality and availability tokens to route and modulate pooled queries from the fused visual representation, enabling target-adaptive conditioning without dense voxel-to-voxel attention. A conditional flow matching objective transports target-channel noise toward the single missing sequence, while four Heun trajectories are averaged to produce the completion. Evaluation across structural, diffusion, and contrast-enhanced MRI in UCSF-PDGM demonstrates consistent whole-brain and tumor-region fidelity across all eight single-target completion tasks, with the strongest performance on structural sequences. Code is available at \mbox{\urlstyle{same}\url{https://github.com/CUHK-AIM-Group/MOSAIC-FLOW}}.
\end{abstract}


\section{Introduction}
Multimodal magnetic resonance imaging (MRI) provides complementary information on brain anatomy, tissue microstructure, perfusion, susceptibility, and contrast enhancement, which is important for characterizing brain tumors clinically \cite{hao2026panda}. However, sequences may be unavailable because of scan time constraints, motion, safety considerations, or heterogeneous clinical protocols. Single missing sequence completion aims to recover the unavailable target from the co registered modalities while preserving modality specific contrast and subtle three dimensional tumor anatomy.

Missing MRI synthesis has been studied using adversarial generation, modality masking, shared representations, and latent generative models \cite{Liu2023MMT,Zhang2025UMM,Zhang2025MISA,Qi2026Unified,PanHao_TaskAdaptive_MICCAISAT2026}. Diffusion based methods further enable flexible synthesis from incomplete multimodal inputs \cite{zhu2025cycle,zhu2025q,pang2026cascaded,Zhong2026FMMDiff,PanHao_nnDiffusion_MICCAISAT2026}, with accelerated sampling strategies reducing inference cost \cite{Nichol2021ImprovedDDPM,Song2021DDIM}. Recent volumetric approaches improve three dimensional synthesis through orientation aware refinement, efficient multimodal fusion, and structural constraints \cite{Wei2026RTFSyn,Zhang2026CSSDiff,Pang2026D3M}, while conditional flow matching provides an alternative continuous generative formulation for controllable MRI synthesis \cite{lipman2023flow,Chang2025ControllableFM,Yazdani2026Flow}. 
Despite these advances, unified completion of arbitrary missing sequences in heterogeneous brain tumor MRI remains challenging. First, a shared model must preserve modality-specific contrast and fine-grained 3D tumor anatomy while adapting to the target modality. Second, simple channel concatenation may obscure modality identity, whereas dense voxel-wise fusion incurs high computational costs. These challenges call for efficient, target-aware integration of complementary multimodal information.

To this end, we propose MOSAIC-Flow, a unified framework for single missing sequence completion in co-registered brain MRI. The model separates modality specific encoding from shared volumetric processing and learns a target conditioned velocity field for one missing sequence at a time. The acquired modalities remain fixed throughout transport, while the missing target is reconstructed using a shared generative model. The contributions are summarized as follows:
\begin{enumerate}
  \item We propose MOSAIC-Flow for unified completion across eight MRI sequences. Modality aware stems preserve sequence identity, while \textit{target conditioned flow matching reconstructs the requested modality without modifying the seven acquired inputs during transport}.
  \item We introduce orthogonal axis refinement to \textit{capture axial, coronal, and sagittal dependencies through identity initialized soft gating while preserving the three dimensional representation at the bottleneck}.
  \item We develop status conditioned prototype fusion, which uses modality and acquisition status tokens to route and modulate pooled visual queries for \textit{target adaptive conditioning without dense voxel level attention}. Comprehensive experiments and matched ablations demonstrate consistent improvements across all eight completion targets.
\end{enumerate}

\section{Methodology}\label{sec:method}
\begin{figure*}[!t]
	\centering
	\includegraphics[width=0.9\textwidth]{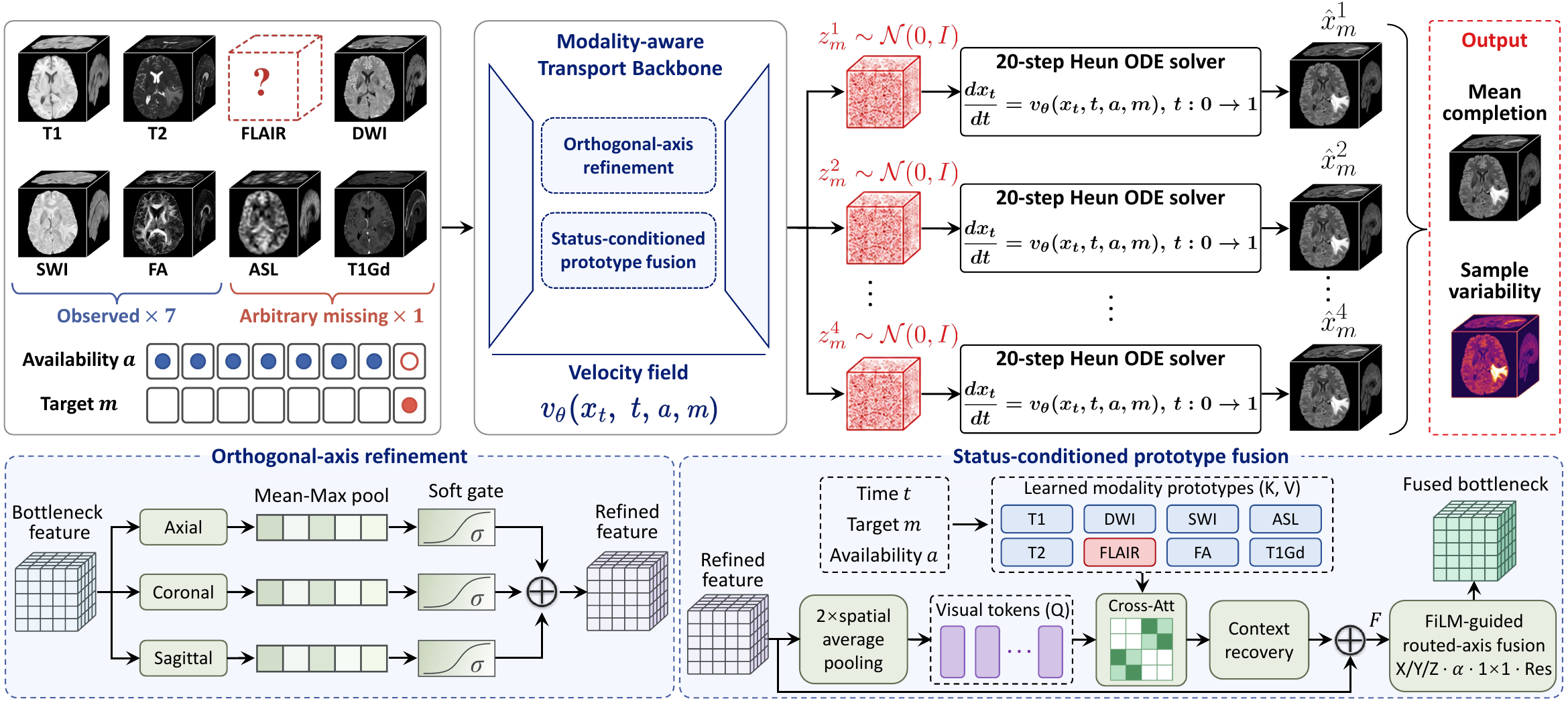}
	\vspace{-10pt}
\caption{\textbf{Overview of MOSAIC-Flow.} Seven acquired modalities, availability vector $a$, and target vector $m$ condition the 3D transport backbone. Orthogonal axis refinement, axial self attention, and status conditioned prototype fusion operate at the bottleneck. Four independent noise samples are integrated using 20 step Heun trajectories and averaged for the final completion.}
\label{fig:method}
\vspace{-10pt}
\end{figure*}
\subsection{Overview}
Figure~\ref{fig:method} illustrates MOSAIC-Flow, comprising modality aware encoding, a shared 3D transport backbone, orthogonal axis refinement, status conditioned prototype fusion, and a shared velocity decoder. The framework reconstructs one requested target modality from the other seven acquired sequences while keeping the observed volumes fixed throughout transport.
Let $X=(x^1,\ldots,x^M)\in\mathbb{R}^{M\times D\times H\times W}$ denote $M=8$ co registered MRI volumes, where $D$, $H$, and $W$ are the spatial dimensions. A one hot vector $m\in\{0,1\}^{M}$ identifies the missing target, and $a=\mathbf{1}-m$ denotes modality availability, so exactly seven sequences are observed. We define $X_{\mathrm{obs}}=a\odot X$, where $\odot$ denotes channel wise multiplication with vectors broadcast over space. For Gaussian noise $Z\sim\mathcal{N}(0,I)$ and time $t\sim\mathcal{U}[0,1]$, the conditional probability path and target velocity are:
\begin{equation}
\begin{aligned}
 X_t &= X_{\mathrm{obs}}+m\odot\big((1-t)Z+tX\big),\\
 U &= m\odot(X-Z),\qquad
 \widehat U_\theta=v_\theta(X_t,t,X_{\mathrm{obs}},a,m),
\end{aligned}
\label{eq:overview}
\end{equation}
where the requested target is interpolated from noise at $t=0$ to the acquired target at $t=1$, while all observed channels remain fixed at $X_{\mathrm{obs}}$ throughout the path.

\subsection{Target Conditioned Flow Matching}
To learn the transport of the requested target while preserving the acquired sequences, we estimate the conditional velocity $U$ in Eq.~\eqref{eq:overview} using a target restricted regression objective. Let $w_i$ denote the spatial weight at voxel $i$, and define $\lVert Y\rVert_w^2=(\sum_i w_iY_i^2)/(\sum_i w_i)$. The flow matching objective~\cite{lipman2023flow} is
\begin{equation}
\mathcal L_{\mathrm{FM}}=
\E_{X,Z,t,m}\!\left[
\left\|m\odot\left(\widehat U_\theta-U\right)\right\|_w^2
\right],
\label{eq:fm}
\end{equation}
where the expectation is taken over training volumes, target choices, noise samples, and interpolation times, and $m$ restricts supervision to the requested target. During inference, the observed channels are reset to $X_{\mathrm{obs}}$ after each predictor and corrector update to prevent numerical drift.

\subsection{Modality Aware Encoding}
To preserve modality identity before shared volumetric processing, we augment the shared representation with modality specific residual branches. For modality $k$, a two layer 3D stem $f_k$ processes the current path volume $x_t^k$, the observed volume $x_{\mathrm{obs}}^k$, and the availability and target indicators $a_k$ and $m_k$, which are broadcast over the spatial grid. Denoting channel concatenation by $[\cdot]$ and concatenation across modalities by $\parallel$, the initial feature map is:
\begin{equation}
\begin{aligned}
F_0={}&F_{\mathrm{shared}}([X_t,X_{\mathrm{obs}},a,m])\\
&+P_0\!\left(
\mathop{\parallel}_{k=1}^{M}
f_k[x_t^k,x_{\mathrm{obs}}^k,a_k,m_k]
\right),
\end{aligned}
\label{eq:stem}
\end{equation}
where $F_{\mathrm{shared}}$ is the shared 3D stem and $P_0$ is a $1\times1\times1$ projection. The projected modality specific features provide a residual correction to the shared representation. Sinusoidal time embeddings together with learned availability and target embeddings subsequently modulate the residual blocks of the four level 3D U-Net. Initializing $P_0$ to zero preserves the shared stem mapping at the start of optimization while allowing modality specific corrections to be learned.

\subsection{Orthogonal Axis Refinement}
To capture complementary dependencies along the three anatomical orientations while preserving the 3D representation, we refine the bottleneck feature $F_b$ along the depth, height, and width axes. For each orientation $p\in\{D,H,W\}$, $\mathrm{AvgMax}_p(\cdot)$ forms a two channel 2D descriptor by average and maximum pooling over the feature channels and axis $p$, thereby preserving the corresponding orthogonal plane. An orientation specific network $g_p$ then predicts a soft gate:
\begin{equation}
\begin{aligned}
G_p &=
2\sigma\left(
g_p\left(\mathrm{AvgMax}_p(F_b)\right)
\right),\\
F_{\mathrm{axis}} &=
\frac{1}{3}
\sum_{p\in\{D,H,W\}}
F_b\odot G_p.
\end{aligned}
\label{eq:axis}
\end{equation}
Each gate is broadcast to the size of $F_b$ before element wise modulation. The final projection of each $g_p$ is initialized to zero, yielding $G_p=1$ at initialization, so the refinement starts as an identity mapping. We subsequently apply axial self attention along the three spatial axes to $F_{\mathrm{axis}}$, producing $F_{\mathrm{att}}$ with extended volumetric context before conditional fusion.

\subsection{Status Conditioned Prototype Fusion}
To adapt the refined visual representation to the requested target and acquisition status without dense spatial cross attention, we introduce status conditioned prototype fusion. We first apply factor two average pooling $P$ to $F_{\mathrm{att}}$ and flatten the result into visual queries $Q$. For modality $k$, a condition token $C_k$ combines a learned modality prototype $e_k^{\mathrm{mod}}$, a binary status embedding $e_k^{\mathrm{status}}$, and the projected global condition embedding $W_e e$, where $e$ encodes time, availability, and target information for conditional fusion:
\begin{equation}
\begin{aligned}
Q&=\mathrm{vec}(P(F_{\mathrm{att}})),\\
C_k&=e_k^{\mathrm{mod}}+e_k^{\mathrm{status}}+W_e e,\\
C&=[C_1;\ldots;C_M],\qquad
\widetilde Q=\mathrm{Attn}(Q,C,C).
\end{aligned}
\label{eq:fusion}
\end{equation}
Here, $Q$ provides the query sequence, while $C$ provides both the key and value sequences for cross attention. The attended tokens are reshaped, interpolated to the bottleneck resolution, and injected into $F_{\mathrm{att}}$ through a zero initialized projection, while their mean context generates channel wise affine modulation and per channel mixture weights for depth, height, and width convolutions. The weighted responses are then projected and added residually, enabling compact condition tokens to guide the visual representation according to modality identity, acquisition status, time, and target without introducing modality specific spatial features.

\subsection{Objective and Inference}\label{subsec:objective}
Let $B$ and $T$ denote the binary brain and tumor masks, and let $\partial T$ denote the tumor boundary. We define the spatial weight as $w=0.05+0.95B+3T+2\partial T$ to emphasize brain anatomy, tumor regions, and lesion boundaries. In addition to the flow matching loss, we estimate the target endpoint as $\widehat X=X_t+(1-t)\widehat U_\theta$ and impose weighted intensity, gradient, and local 3D structural constraints. The complete objective is:
\begin{equation}
\mathcal L=\mathcal L_{\mathrm{FM}}
+0.20\mathcal L_{1}^{w}
+0.03\mathcal L_{\nabla}^{w}
+0.08\mathcal L_{\mathrm{SSIM3D}}^{w},
\label{eq:loss}
\end{equation}
where $\mathcal L_{1}^{w}$, $\mathcal L_{\nabla}^{w}$, and $\mathcal L_{\mathrm{SSIM3D}}^{w}$ constrain target intensity, first order spatial gradients, and local 3D structure, respectively. During inference, $K=4$ independent noise volumes are integrated from $t=0$ to $t=1$ using a 20 step Heun solver, with the observed channels restored after each update. The final completion is the sample mean $x_m=K^{-1}\sum_{s=1}^{K}\widehat{x}_m^{(s)}$.

\subsection{Implementation}
\noindent\textbf{Dataset.} We evaluated MOSAIC-Flow on UCSF-PDGM~\cite{Calabrese2022UCSF}, comprising 501 scans from 495 subjects with T1, T2, FLAIR, DWI, SWI, HARDI derived FA, ASL, and T1Gd. Intensities were clipped at the 0.5th--99.5th percentiles, and reversibly cropped to $192\times208\times152$ without excluding tumors. 

\noindent\textbf{Optimization.} MOSAIC-Flow used a base width of 32, channel multipliers $(1,2,4,8)$, two residual blocks per level, a 512 dimensional condition embedding, and eight attention heads. The model was trained on NVIDIA H200 GPUs using AdamW with a learning rate of $5\times10^{-5}$, weight decay of $10^{-5}$, effective batch size of four, BF16 and TF32 computation, and an EMA rate of 0.9999.

\section{Experiments}
\subsection{Completion Across All Target Modalities}
\begin{figure}[!t]
\centering
\includegraphics[width=\columnwidth]{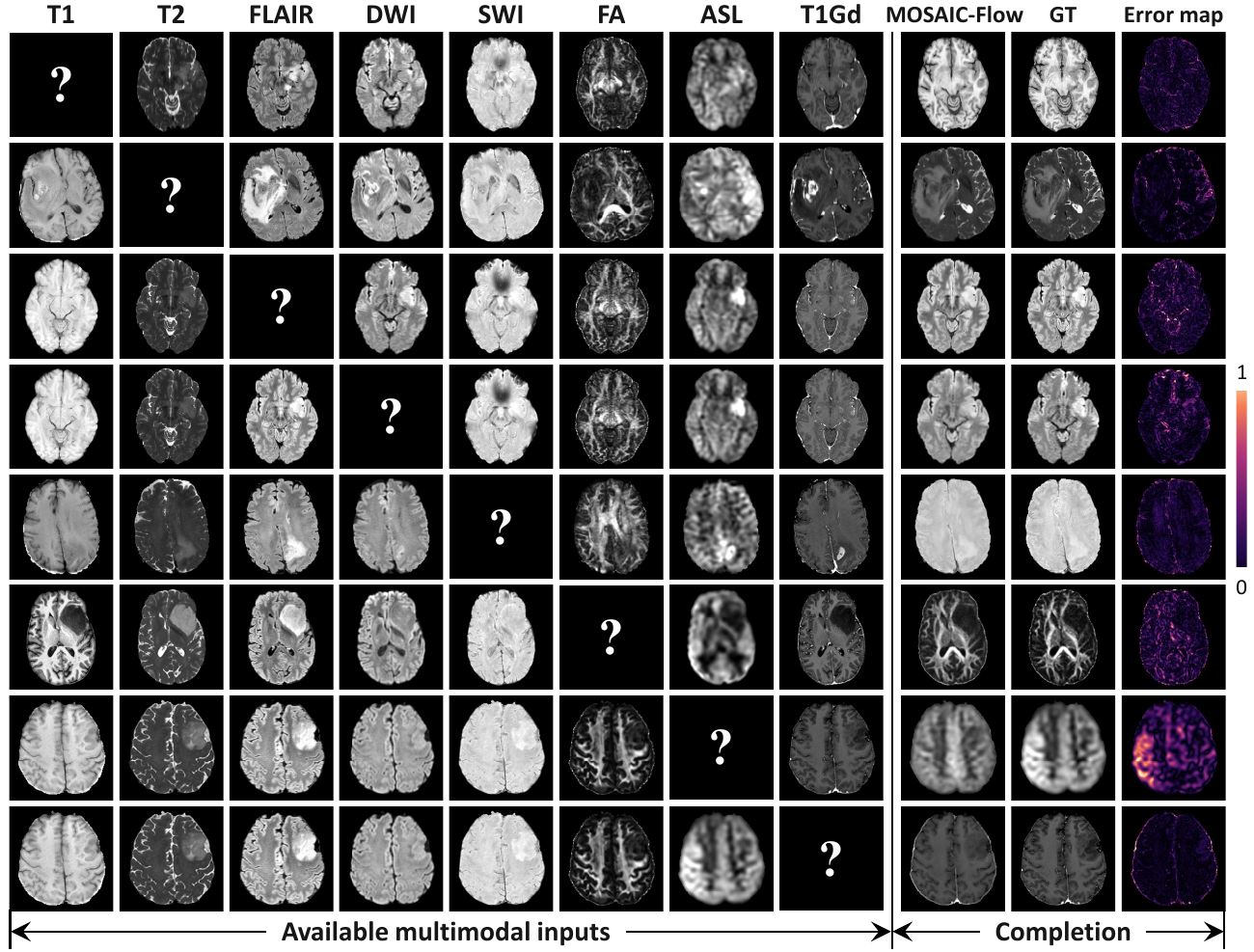}
\vspace{-22pt}
\caption{\textbf{Completion across eight target modalities.} Each row shows seven available inputs, the missing target, MOSAIC-Flow completion, ground truth, and absolute error.}
\label{fig:completion}
\vspace{-1pt}
\end{figure}

We evaluated completion across all eight target modalities in UCSF-PDGM. Comparisons included Flow Matching~\cite{lipman2023flow} and VolSynth~\cite{bottani2024contrast}, with MAE, PSNR, and MS-SSIM evaluated in both brain and tumor regions. As shown in Fig.~\ref{fig:completion}, MOSAIC-Flow reconstructs the distinct appearance of all eight target modalities while preserving overall brain anatomy in the representative case. The error maps further show target dependent residual patterns, with larger errors observed for ASL, likely reflecting its lower signal to noise ratio and greater physiological variability.

\begin{figure}[!t]
	\centering
	\includegraphics[width=0.98\columnwidth]{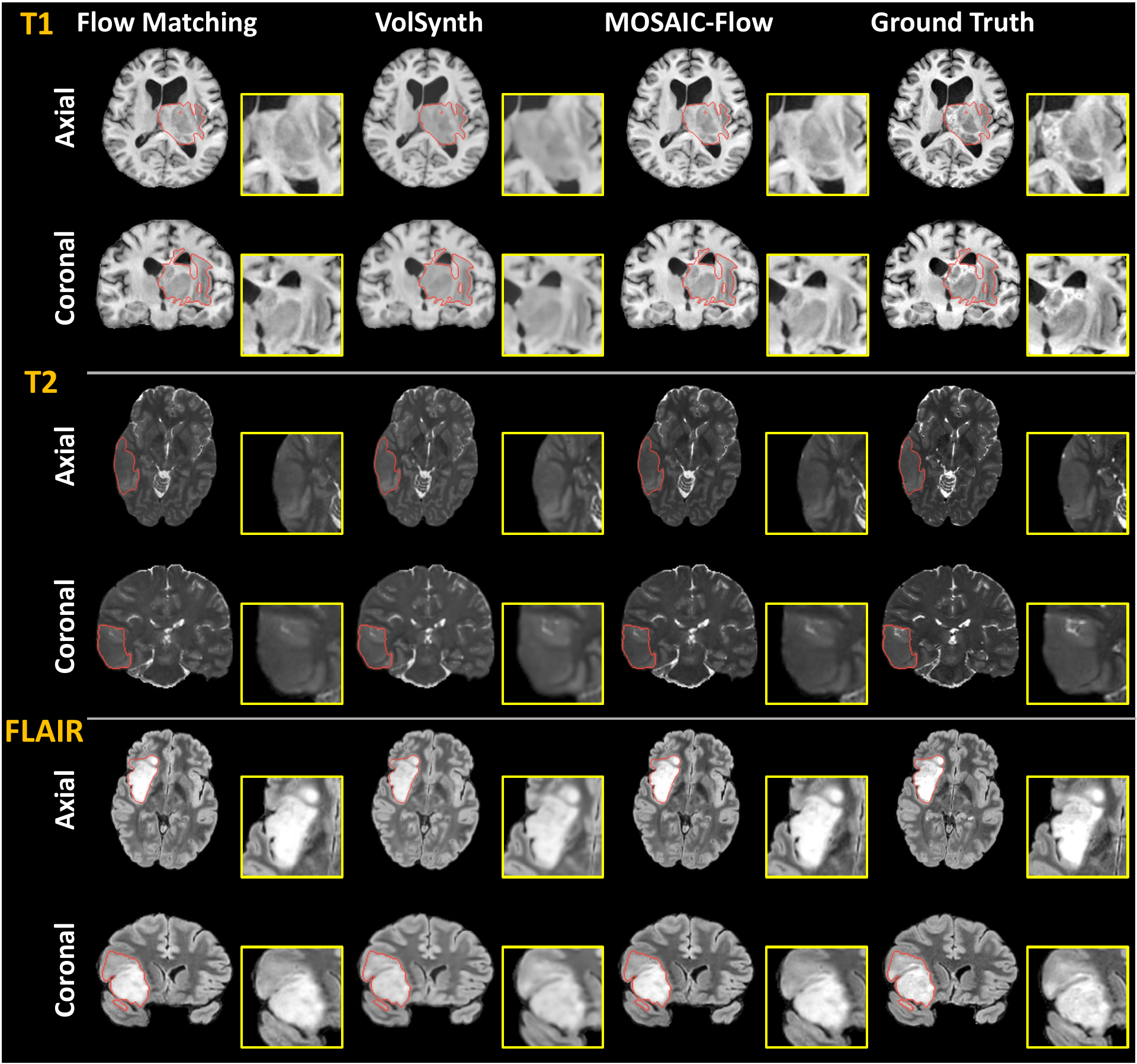}
	\vspace{-10pt}
	\caption{\textbf{Tumor-focused qualitative comparison.} Matched axial and coronal views are shown for Flow Matching, VolSynth, MOSAIC-Flow, and GT. Red contours mark the tumor region, and yellow boxes show magnified local anatomy.}
	\label{fig:comparison}
	\vspace{-5pt}
\end{figure}

\subsection{Anatomical Fidelity in Tumor Regions}
To assess anatomical fidelity in tumor regions, Fig.~\ref{fig:comparison} presents tumor-focused comparisons for T1, T2, and FLAIR using axial and coronal views. Across all targets, MOSAIC-Flow better preserves lesion boundaries and local contrast than Flow Matching and VolSynth, showing closer agreement with ground truth. This highlights its ability to preserve fine-grained tumor anatomy across structural contrasts.

\subsection{Quantitative Performance Across MRI Sequences}
\begin{figure}[!t]
\centering
\includegraphics[width=1\columnwidth]{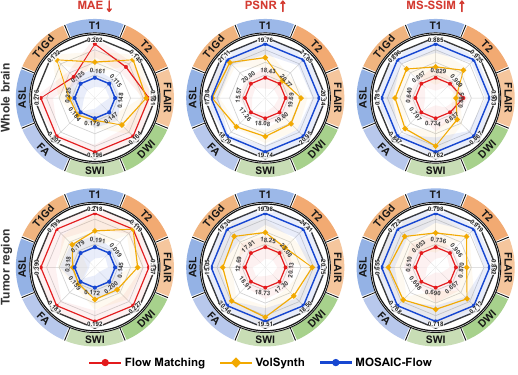}
\vspace{-20pt}
\caption{\textbf{Quantitative comparison across eight target modalities.} Radar plots show mean performance with shaded Mean $\pm$ SD for whole-brain and tumor-region evaluation. Each spoke represents one target modality. Lower MAE and higher PSNR and MS-SSIM indicate better performance.}
\label{fig:quantitative}
\vspace{-5pt}
\end{figure}

To quantify completion performance across MRI sequences, Fig.~\ref{fig:quantitative} reports mean MAE, PSNR, and MS-SSIM across all eight targets for whole-brain and tumor regions, with shaded bands denoting standard deviation (SD). MOSAIC-Flow achieves the lowest MAE and highest PSNR and MS-SSIM for every target in both regions, consistently outperforming Flow Matching and VolSynth across all sequence types. These results demonstrate robust and consistent completion across the full eight-modality MRI protocol.

\subsection{Ablation Studies}
To evaluate each proposed module, we remove orthogonal-axis refinement, status-conditioned prototype fusion, or both under identical settings. Table~\ref{tab:ablation} shows that removing either module degrades all metrics in both brain and tumor regions, while removing both yields the largest degradation, supporting the contribution of each module.
\begin{table}[!t]
\centering
\caption{\textbf{Component ablation across all eight targets.} Values are averaged over the same test split. Bold denotes the best value for each region and metric.}
\label{tab:ablation}
\resizebox{\columnwidth}{!}{%
\begin{tabular}{llccc}
\toprule
Region & Configuration & MAE $\downarrow$ & PSNR $\uparrow$ & MS-SSIM $\uparrow$ \\
\midrule
Brain & w/o axis & 0.2149 & 17.3891 & 0.7331 \\
 & w/o fusion & 0.2145 & 17.5029 & 0.7605 \\
 & w/o both & 0.2258 & 16.7493 & 0.7290 \\
 & Full Model & \textbf{0.1605} & \textbf{19.9275} & \textbf{0.8583} \\
\addlinespace[1pt]
Tumor & w/o axis & 0.2318 & 17.5129 & 0.6423 \\
 & w/o fusion & 0.2332 & 17.4701 & 0.6595 \\
 & w/o both & 0.2470 & 16.3239 & 0.6316 \\
 & Full Model & \textbf{0.1829} & \textbf{19.7456} & \textbf{0.7731} \\
\bottomrule
\end{tabular}}
\end{table}

\section{Conclusion}
We present MOSAIC-Flow, a unified framework for single-missing-sequence completion across eight brain MRI modalities. By combining modality-aware encoding, orthogonal-axis refinement, status-conditioned prototype fusion, and conditional flow matching, MOSAIC-Flow improves whole-brain and tumor-region reconstruction across heterogeneous MRI contrasts while preserving anatomical fidelity. Extension to multiple missing sequences remains future work.

\section{Compliance with ethical standards}
This study used the publicly available, de-identified UCSF-PDGM dataset in accordance with its data-use terms. No additional human participant data were collected.

\section{Acknowledgments}
This work was supported by NSFC/RGC Joint Research Scheme N\_CUHK4126/25 and Innovation and Technology Fund Mainland-Hong Kong Joint Funding Scheme MHP/173/24. The authors declare no conflicts of interest.

\raggedbottom
\bibliographystyle{IEEEbib}
\bibliography{paper}
\end{document}